\documentclass[conference]{IEEEtran}
\ifCLASSOPTIONcompsoc
  \usepackage[nocompress]{cite}
\else
  \usepackage{cite}
\fi
\ifCLASSINFOpdf
\else
\fi
\usepackage{url}

\usepackage{graphicx}
\usepackage{amsmath}
\usepackage{amssymb}
\usepackage{cleveref}
\usepackage[table,xcdraw]{xcolor}
\usepackage{todonotes}
\usepackage{color}
\usepackage{listings}
\usepackage{caption}
\usepackage{float}
\usepackage{subcaption}

\newcounter{nalg} 
\renewcommand{\thenalg}{\arabic{nalg}} 
\DeclareCaptionLabelFormat{algocaption}{Algorithm \thenalg} 
\lstnewenvironment{algorithm}[1][] 
{   
    \refstepcounter{nalg} 
    \captionsetup{labelformat=algocaption,labelsep=colon} 
    \lstset{ 
        mathescape=true,
        frame=tB,
        numbers=left, 
        numberstyle=\tiny,
        basicstyle=\scriptsize, 
        keywordstyle=\color{black}\bfseries\em,
        keywords={,do, return, function, in, if, elif, else, and, or, not, forall, foreach, while, begin, end, for, break, } 
        numbers=left,
        escapechar=§,
        xleftmargin=.04\textwidth,
        #1 
    }
}
{}
\crefname{lstlisting}{Algorithm}{Algorithms}
\Crefname{lstlisting}{Algorithm}{Algorithms}

\begin{document}
%
\title{Event Abstraction for Enterprise Collaboration Systems to Support Social Process Mining}

\author{\IEEEauthorblockN{Jonas Blatt}
\IEEEauthorblockA{University of Koblenz\\
Koblenz, Germany\\
Email: jonasblatt@uni-koblenz.de}
\and
\IEEEauthorblockN{Patrick Delfmann}
\IEEEauthorblockA{University of Koblenz\\
Koblenz, Germany\\
Email: delfmann@uni-koblenz.de}
\and
\IEEEauthorblockN{Petra Schubert}
\IEEEauthorblockA{University of Koblenz\\
Koblenz, Germany\\
Email: schubert@uni-koblenz.de}}


%


\maketitle

\begin{abstract}
One aim of Process Mining (PM) is the discovery of process models from event logs of information systems.
PM has been successfully applied to process-oriented enterprise systems but is less suited for communication- and document-oriented Enterprise Collaboration Systems (ECS).
ECS event logs are very fine-granular and PM applied to their logs results in spaghetti models. 
A common solution for this is event abstraction, i.e., converting low-level logs into more abstract high-level logs before running discovery algorithms.
ECS logs come with special characteristics that have so far not been fully addressed by existing event abstraction approaches.
We aim to close this gap with a tailored ECS event abstraction (ECSEA) approach that trains a model by comparing recorded actual user activities (high-level traces) with the system-generated low-level traces (extracted from the ECS).
The model allows us to automatically convert future low-level traces into an abstracted high-level log that can be used for PM. 
Our evaluation shows that the algorithm produces accurate results.
ECSEA is a preprocessing method that is essential for the interpretation of collaborative work activity in ECS, which we call \textit{Social Process Mining}.
\end{abstract}


%
\IEEEpeerreviewmaketitle

\def\mi#1{\mathit{#1}}
\def\mc#1{\mathcal{#1}}
\def\seq#1{\langle #1 \rangle}


\newtheorem{example}{Example}
\newtheorem{definition}{Definition}
\section{Introduction}\label{sec:introduction}
Over the past 20 years, Process Mining (PM) has gained significance in research and has also become an essential part of business process management efforts in companies in the last decade  \cite{VomBrocke2021}. 
One of the main tasks of PM is the discovery of process models from business software event logs \cite{VanDerAalst2022}. 
Typically, PM is applied to transaction-processing information systems such as workflow management, ERP, or supply chain management systems \cite{VanderAalst2007}.

Recently, companies increasingly implemented Enterprise Collaboration Systems (ECS). 
They contain features known from Social Media and groupware, support computer-mediated collaboration and communication, and have become essential components in the digital workplace in many companies \cite{Wehner2017,Williams2018}.
Although PM can potentially reveal essential insights for understanding collaboration and thus shed more light on how people collaborate, research and applications of PM in collaboration systems are scarce. 
One reason for this is that, in contrast to the business software types mentioned above, an ECS supports ad hoc collaboration and works on documents through specific features such as creating blog posts or wiki pages and editing, replying, or following them. 
Work supported by such activities is significantly less structured compared to work in, for example, ERP systems.

In an early exploratory study, van der Aalst applied PM algorithms to different collaboration systems and discovered several challenges. 
One major challenge was that the logs of collaboration systems are too fine-granular \cite{VanderAalst2005}, also referred to as low-level (LL) logs \cite{Gunther2006}. 
Applying PM to such logs results in so-called spaghetti-models that are overly complex and hard to interpret \cite{Rebmann2022}.
In preparation for this research, we have applied state-of-the-art PM algorithms to event logs from HCL Connections, one of the market-leading integrated ECS. 
We can confirm that, among many others, the outlined challenges persist (see \Cref{sec:data_unterstanding}). 
Thus, with state-of-the-art methods, information gained from applying PM to ECS logs is limited.

The research direction Social Process Mining (SPM) aims to identify collaboration patterns in ECS.
In particular, the interpretation of collaborative work activity in ECS should be enabled by using methods from the PM domain.
The coarse SPM process contains steps for event abstraction, case identification, process discovery, and frequent subgraph mining for pattern detection.
Other than the field of Social Network Analysis, which focuses on social network mining, determination of network metrics (centrality etc.), and community detection \cite{tabassumSocialNetworkAnalysis2018}, SPM is about how users of ECS ``move through the system'' i.e., which functions they typically execute in which order and how they collaborate on social content.

This paper focuses on the LL log (LLL) of collaboration systems and develops a novel approach for event abstraction (EA) as an important preprocessing step in SPM. 
The granularity of event logs is a common problem in PM \cite{Bose2009,Gunther2009,Andrews2018,Fazzinga2018,DeLeoni2020,Dumas2018}, which is typically addressed through EA \cite{VanZelst2021}. 
EA helps with the translation of ``(multiple) instances of fine-granular events into instances of coarse-granular events'' \cite[p.\,724]{VanZelst2021}. 
The aim is to bring the event log to a level where the events represent interpretable activities.
Our EA approach addresses the specific characteristics of ECS event logs and can generate a high-level log (HLL) from the LLL of collaboration software. 
\textit{ECS Event Abstraction} (ECSEA) is a specific type of EA.
In a nutshell, the ECSEA algorithm trains a model by comparing HLL traces with the related LLL traces. 
The LLL is extracted from the ECS log database, and the HLL is built by observing (recording) the ECS use for a defined period of time using a click path observer (for details, we refer to \Cref{subsec:gathering_high_level_log}). 
For organizational and/or legal reasons, it is not always possible to permanently observe an ECS to obtain a HLL. Therefore, the idea is to observe an ECS in a controlled environment for a certain period of time, and in this way, to obtain a HLL generated by click path observation. 
The trained ECSEA model can then be used to transform historic or future LLL into (previously unknown) HLLs.
This is possible because it can be assumed that the LLL behavior of the ECS of the same instance will not change over time.
For other instances of the same system type or for systems of a different vendor, the observation can be repeated if the system configuration is too different or customized.

In the remainder, \Cref{sec:data_unterstanding} introduces ECS event logs and outlines their characteristics.
\Cref{sec:literature_review} reviews EA approaches in the context of ECS logs. 
\Cref{sec:ecs_event_abstraction} presents the ECSEA approach, which is evaluated in \Cref{sec:evaluation}. 
We conclude with a discussion in \Cref{sec:discussion_outlook}.

\section{Data Understanding: ECS Logs}\label{sec:data_unterstanding}
In this section, the log characteristics of ECS are described.
First, we demonstrate an event log of HCL Connections as an example (\ref{subsec:ECS-event-log-example}).
Then, we outline its special characteristics
(\ref{subsec:ecs_event_log_characteristics}).
\subsection{ECS Event Log Example}\label{subsec:ECS-event-log-example}
ECS commonly use relational databases to store event logs. 
\Cref{tab:ecs_event_log_excerpt} shows an excerpt of a corresponding event log table of HCL Connections.
\newcommand{\hcell}[1]{\cellcolor[HTML]{656565}{\color[HTML]{FFFFFF} \texttt{#1}}}
\begin{table}[h!]
\caption{ECS LL log excerpt}\label{tab:ecs_event_log_excerpt}
\scriptsize
\resizebox{1.0\columnwidth}{!}{
\begin{tabular}{|l|l|l|l|l|l|l}
\cline{1-6}
\hcell{ID} & \hcell{USER\_UUID} & \hcell{ITEM\_UUID} & \hcell{C\_ID} & \hcell{EVENT\_TS} & \hcell{EVENT\_NAME} \\
\cline{1-6}
104 & 33d689884-ef3... & 072adbbd-715... & 1337 & 2021-11-11 10:47:16 & file.file.created &  \\ \cline{1-6}
105 & 33d689884-ef3... & 072adbbd-715... & 1337 & 2021-11-11 10:47:17 & file.collection.file.added &  \\ \cline{1-6}
106 & 84f5d4ae2-434... & 08441212-343... & 1337 & 2021-11-11 10:47:18 & wiki.page.created &  \\ \cline{1-6}
107 & 33d689884-ef3... & 072adbbd-715... & 1337 & 2021-11-11 10:47:19 & files.file.notification.set &  \\ \cline{1-6}
108 & 84f5d4ae2-434... & 08441212-343... & 1337 & 2021-11-11 10:47:20 & wiki.page.follow &  \\ \cline{1-6}
109 & 33d689884-ef3... & - & 1337 & 2021-11-11 10:48:18 & community.visit &  \\ \cline{1-6}
110 & 65788ec84-a12... & 08441212-343... & 1337 & 2021-11-11 10:51:11 & wiki.page.updated &  \\ \cline{1-6}
111 & 65788ec84-a12... & 08441212-343... & 1337 & 2021-11-11 10:51:12 & wiki.page.tag.added &  \\ \cline{1-6}
112 & 33d689884-ef3... & 08441212-343... & 1337 & 2021-11-11 10:56:44 & wiki.page.updated &  \\ \cline{1-6}
\end{tabular}
}%
\end{table}

In total, this particular database table contains 16 attributes. 
For brevity, we reduced the example to the six most essential attributes. 
Each event record is identified by a unique \texttt{ID}. 
The log documents the user who initiated the event (\texttt{USER\_UUID}), the modified or created content item (\texttt{ITEM\_UUID}), the workspace in which the event occurred (\texttt{C\_ID}), the timestamp (\texttt{EVENT\_TS}) and an activity (\texttt{EVENT\_NAME}).
The IDs in the event log can be resolved to human-readable descriptors using additional dimension tables. 
More details about ECS logs can be found in related work on social collaboration analytics \cite{Schwade2021}.

The log excerpt in \Cref{tab:ecs_event_log_excerpt} documents that a user was uploading a file while another user was creating a wiki page at the same time.
Later, the first user updated the file and created a comment to that file.
This small example illustrates some of the challenges for the interpretation of user activity when using the LLL of HCL Connections. 
Although the first user carried out only one activity (uploading a file), the log recorded two additional events (adding the file to a collection and setting a notification).
For interpretation in the context of PM, only the first event (104) is relevant as this event represents the user’s intended action.
The follow-up events (105, 107) are tasks automatically performed by the system (and they may occur in changing orders). 
However, they are associated with the user.
Such event sequences where multiple LL events are recorded for one actual activity lead to spaghetti models.
Moreover, LL events are not distinct to only one specific HL activity but can be part of different HL activities (example see below). 
Furthermore, HCL Connections contains more than 250 different LL activities. 
This is also an indicator for the need to reduce the number of activities to the ones that are useful for the interpretation of user activity.
In ECS event logs, certain events may have no meaning at all for later interpretation.
For instance, the activity \textit{community.visit} (e.g.,\,event 109) occurs over and over again, while it does not contribute to the understanding of the actual process. 
Given this example event log as the first impression of an ECS event log, the following subsection shows problems and challenges that may occur in such event logs.
\subsection{ECS Event Log Characteristics}\label{subsec:ecs_event_log_characteristics}
As demonstrated in the example above, multiple LL events can belong to one HL activity. 
We also observed that their occurrence might differ depending on other events or previous user activities (e.g.,\,some activities are only triggered once during a session).
The following list summarizes the characteristics and challenges with ECS LLLs in the context of PM. 
While single mentions of some of these challenges were found in related work \cite{VanderAalst2005,naderipour_mining_2011,Poncin2011,Schwade2019} and occur in traditional event logs, ECS event logs contain all of the following challenges at the same time:
\begin{enumerate}
    \item[C1] \textbf{Multiple LL events}: Particular HL events may result in multiple LL events. 
    Thus, a HL activity may be expressed by multiple LL activities. 
    Furthermore, a particular HL activity may be expressed by different sets of LL activities.
    \textit{Example: The LL events 104, 105 and 107 should be combined into one HL event.}
    \item[C2] \textbf{Overlapping events}: Multiple LL events that are related to two (or more) HL events may overlap temporally.
    Moreover, the interval between the time-stamps of multiple LL events that are triggered by an activity is not always the same. 
    \textit{Example: The LL events 104, 105 and 107 (HL event A) and the LL events 106 and 108 (HL event B) are overlapping.}
    \item[C3] \textbf{Different LL event ordering}: The ordering of the LL events might vary slightly. This is because the underlying system performs tasks independently so that other activities may be logged first.
    \textit{Example: The LL sequence $\seq{104, 105, 107}$ may also occur as $\seq{104, 107, 105}$, while both cases belong to the same HL event.}
    \item[C4] \textbf{Multiple LL to HL activity mappings}: 
    A LL activity may be triggered by multiple HL activities.
    Thus, it is possible that a LL activity is part of more than one mapping.
    This characteristic represents an m:n mapping.
    \textit{Example: The HL activity \textbf{gws.wiki.wikiarticle.tag.created} is composed of the LL activities \textbf{wiki.page.updated} and \textbf{wiki.page.tag.added} (events 110 \& 111). However, the LL activity \textbf{wiki.page.updated} (event 112) is also part of the HL activity \textbf{gws.wiki.wikiarticle.updated}. }
    \item[C5] \textbf{Ghost activities}: There may exist LL events with no related HL activity.
    \textit{Example: Event 109 named \textbf{community.visit} has no related HL activity.}
\end{enumerate}
Addressing these challenges requires an EA technique that can handle the outlined characteristics of ECS LLLs and is capable of generating interpretable HLLs. 
The outlined challenges serve as requirements for ECSEA. 
\Cref{sec:literature_review} is dedicated to a review of existing approaches for EA.

\section{Related Work}\label{sec:literature_review}
We investigated existing EA approaches with a focus on ECS log challenges (\Cref{subsec:ecs_event_log_characteristics}).
For this, a backward search based on a recent extensive seminal literature review on EA by van Zelst et al. was performed \cite{VanZelst2021}.
In an additional forward search, five relevant publications on EA were added (Andrews et al.\ (2018) \cite{Andrews2018}, Klessascheck et al.\ (2021) \cite{Klessascheck2021}, Li et al.\ (2021) \cite{Li2021}, Rebmann et al.\ (2022) \cite{Rebmann2022} and Faziinga et al.\ (2022) \cite{Fazzinga2022}). In total, we reviewed 25 publications. 
\Cref{tab:literature-review-requirements} shows the results of the analysis of the retrieved references in a concept matrix containing the challenges as dimensions (\textbf{C1}-\textbf{C5}). 
For each of the existing approaches in the literature, we investigated if the approach works without manual preprocessing based on a-priori domain knowledge, which is shown in \textbf{C6}.
\newcommand{\eaa}[9]{\cite{#1} & #2 ({#3}) & #4 & #5 & #6 & #7 & #8 & #9 \\ \hline}
\newcommand{\eaaT}[1]{\footnotesize\rotatebox{-90}{\textbf{#1}\hspace{2mm}}}
\newcommand{\etal}{et al.}
\begin{table}[h!]
    \centering
    \scriptsize
    \vspace*{-0.1cm}
    \caption{Related Work covering the ECS log characteristics}
    \resizebox{.5\textwidth}{!}{%
    \begin{tabular}{|l|l||c|c|c|c|c|c|}
        \hline
        \textbf{Ref} &
        \textbf{Author \& Year} &
        C1 &
        C2 &
        C3 &
        C4 &
        C5 &
        C6
        \\
        \hline
        \hline
\eaa{Bose2009}{Bose \& van der Aalst}{2009}{\checkmark}{$\circ$}{$\times$}{$\times$}{$\times$}{$\times$}
\eaa{Gunther2009}{Günther \etal}{2009}{\checkmark}{\checkmark}{\checkmark}{$\times$}{$\times$}{$\times$}
\eaa{Baier2013}{Baier \& Mendling}{2013}{\checkmark}{\checkmark}{\checkmark}{\checkmark}{\checkmark}{$\times$}
\eaa{Ferreira2013}{Ferreira \etal}{2013}{\checkmark}{$\times$}{\checkmark}{$\times$}{$\times$}{$\times$}
\eaa{Folino2015}{Folino \etal}{2015}{\checkmark}{$\times$}{\checkmark}{$\times$}{$\times$}{$\times$}
\eaa{Mannhardt2016}{Mannhardt \etal}{2016}{\checkmark}{\checkmark}{\checkmark}{\checkmark}{\checkmark}{$\times$}
\eaa{Senderovich2016}{Senderovich \etal}{2016}{\checkmark}{\checkmark}{\checkmark}{\checkmark}{\checkmark}{$\times$}
\eaa{VanEck2016}{van Eck \etal}{2016}{\checkmark}{$\times$}{\checkmark}{\checkmark}{$\times$}{$\times$}
\eaa{Begicheva2017}{Begicheva \& Lomazov}{2017}{\checkmark}{\checkmark}{\checkmark}{$\times$}{$\times$}{$\times$}
\eaa{Mannhardt2017a}{Mannhardt \etal}{2017}{\checkmark}{\checkmark}{\checkmark}{\checkmark}{\checkmark}{$\times$}
\eaa{Tax2017}{Tax \etal}{2017}{\checkmark}{$\times$}{\checkmark}{\checkmark}{$\circ$}{$\circ$}
\eaa{Alharbi2018}{Alharbi \etal}{2018}{\checkmark}{$\times$}{\checkmark}{\checkmark}{$\times$}{$\times$}
\eaa{Andrews2018}{Andrews \etal}{2018}{\checkmark}{\checkmark}{\checkmark}{$\times$}{$\times$}{$\times$}
\eaa{Baier2018}{Baier \etal}{2018}{\checkmark}{\checkmark}{\checkmark}{\checkmark}{$\times$}{$\times$}
\eaa{Bernard2018}{Bernard \& Andristos}{2018}{\checkmark}{\checkmark}{\checkmark}{$\times$}{$\times$}{$\times$}
\eaa{Fazzinga2018}{Faszzinga \etal}{2018}{\checkmark}{\checkmark}{\checkmark}{\checkmark}{$\times$}{$\times$}
\eaa{Mannhardt2018}{Mannhardt \etal}{2018}{\checkmark}{\checkmark}{\checkmark}{\checkmark}{\checkmark}{$\times$}
\eaa{SanchezCharles2018}{Sánchez-Charles \etal}{2018}{\checkmark}{\checkmark}{\checkmark}{\checkmark}{$\times$}{$\times$}
\eaa{Leonardi2019}{Leonardi \etal}{2019}{\checkmark}{\checkmark}{\checkmark}{$\times$}{$\times$}{$\times$}
\eaa{Tello2019}{Tello \etal}{2019}{\checkmark}{$\times$}{\checkmark}{\checkmark}{$\times$}{$\times$}
\eaa{DeLeoni2020}{De Leoni \& Dündar}{2020}{\checkmark}{$\times$}{\checkmark}{\checkmark}{$\times$}{$\times$}
\eaa{Klessascheck2021}{Klessascheck \etal}{2021}{$\times$}{$\times$}{$\times$}{$\times$}{\checkmark}{$\times$}
\eaa{Li2021}{Li \etal}{2021}{\checkmark}{\checkmark}{\checkmark}{$\times$}{$\times$}{$\times$}
\eaa{Rebmann2022}{Rebmann \etal}{2022}{\checkmark}{\checkmark}{\checkmark}{\checkmark}{$\times$}{$\times$}
\eaa{Fazzinga2022}{Fazzinga \etal}{2022}{\checkmark}{\checkmark}{\checkmark}{\checkmark}{\checkmark}{$\times$}
\multicolumn{2}{|c||}{\textbf{This work}} & \checkmark & \checkmark & \checkmark & \checkmark & \checkmark & \checkmark \\
        \hline
    \end{tabular}
    }%
    \\[0.2cm]
    {\scriptsize
    \begin{tabular}{lllll}
    & \underline{Challenges} &&&\\
    C1 & Multiple low-level events & \checkmark & = & Meets the challenge\\
    C2 & Overlapping events & $\circ$ & = & Partly meets the challenge \\
    C3 & Different low-level event ordering & $\times$ & = & Does not meet the challenge\\
    C4 & Multiple low-level to high-level &&&\\
      & activity mappings &&&\\
    C5 & Ghost activities &&&\\
    C6 & Works without manual preprocessing&&&\\
      & based on a-priori domain knowledge &&&
    \end{tabular}
    }%
    \label{tab:literature-review-requirements}
\end{table}
\noindent

The literature provides many EA approaches. 
Some of them are tailored for a specific context (e.g.,\, continuous sensor data \cite{Senderovich2016,VanEck2016,Tax2017}, or natural language processing \cite{SanchezCharles2018}), and others can be applied in a general context.
Some approaches need existing process models (representing the process at a HL view) and others a-priory knowledge as input to work properly. 
However, in our context, such a-priory knowledge is not available, and thus these approaches cannot be applied.
Some of the approaches indeed address parts of the characteristics of ECS event logs. 
In the following, we will discuss them and elaborate on why these approaches do not provide a solution in our context either.

Baier and Mendling \cite{Baier2013} introduce a mapping approach that assigns events to activity classes, where a 1:1 mapping is created. Events with no mappings are removed. 
The activity classes are merged according to their context and predefined border conditions with a tree-based incremental clustering algorithm. 
With this approach, a lot of manual effort is required to define such mappings, event context conditions, and border rules.
Furthermore, the naming of the activity classes is straightforward and depends on the correct definition of the context conditions.
Such an approach is unsuitable in a context where no domain expert can define these mappings.

Senderovich et al.\ \cite{Senderovich2016} introduce a knowledge-driven approach with two phases. 
First, an interaction set is defined by creating interactions by selecting, grouping, and filtering LL events. 
Next, based on an optimal matching problem, the correspondences between the interactions and further process knowledge are computed what finally creates the event log.
Again, a deep preliminary understanding of the process is required to define the interaction set.

The work by Mannhardt et al.\ \cite{Mannhardt2017a,Mannhardt2018} introduces similar approaches, where the basic idea is that the LL behavior of activities is expressed in activity patterns. 
These patterns are composed in an abstraction model that is used for the construction of an alignment between events and process steps, which is then used to build the abstracted event log.
As the activity patterns need to be defined (even automatically with Local Process Model discovery \cite{Mannhardt2017a}) and named with the corresponding HL activity, domain knowledge for grouping LL events and HL activities is necessary. 

Fazzinga et al.\ \cite{Fazzinga2022} use a pre-defined list of LL to HL mappings and a pre-defined declarative process model to compute a live interpretation of the current HL activity with augmented reasoning. 
Again, this algorithm requires a pre-defined mapping and a declarative process model.

The approach that is most similar to our approach was developed by Tax et al.\ \cite{Tax2017}, which is based on Conditional Random Fields (CRFs). 
The training data is an annotated event log where each event has a label that refers to the HL activity.
The software that is built on this approach computes a feature space based on all event attributes provided by the log and trains the CRFs, which can transform other unlabeled traces afterwards.
The problem is that this approach is not able to detect overlapping HL activities (represented as interleaving LL events).
Furthermore, we cannot annotate ECS LL events with the real HL activity.
Instead, we are only able to observe the HL sequence of activities where we do not have a relationship between the LL events and the recorded HL events (see \Cref{subsec:gathering_high_level_log}).

Concluding, none of the existing EA approaches are suitable for ECS event logs.
Thus, a new EA approach is required that \textit{a)} can address the outlined challenges, and \textit{b)} does not depend on a-priory domain knowledge. 
In the next section, we present our EA approach that addresses these challenges.
\section{ECS Event Abstraction (ECSEA)}\label{sec:ecs_event_abstraction}
Our EA approach is based on supervised machine learning. 
It learns a model by simultaneously investigating and comparing the observed HL traces with related LL traces in the training phase.
The collection of the HL traces is shortly outlined in \Cref{subsec:gathering_high_level_log}.
After a sufficient amount of HL and LL traces have been recorded, these traces are used to train the model in the first phase (training \& evaluation). 
This procedure is described in \Cref{subsec:ecs_event_abstraction_training}. 
Finally, \Cref{subsec:ecs_event_abstraction_application} introduces the second phase (application), where the model is applied to convert a LL trace into a HL trace.
 \Cref{def:basics1,def:basics2,def:basics3,def:basics4} define preliminary terms for later use in this section.

\begin{definition}\label{def:basics1}
    \textbf{Activity}:
    Let $ \mc{A} $ be the universe of activities.
    Let $ \mc{A}_l \subseteq \mc{A} $ be the set of LL activities and $ \mc{A}_h \subseteq \mc{A} $ be the set of HL activities.
    There exist no overlaps between these two sets so that $ \mc{A}_l \cap \mc{A}_h = \emptyset $
\end{definition}
\begin{definition}\label{def:basics2} 
    \textbf{Event}:
    Let $ \mc{E} $ be the universe of events.
    Let $ \Gamma $ be the universe of attribute names.
    We define $ \#_\gamma(e) $ as a function that assigns an attribute value for the attribute name $ \gamma \in \Gamma $ to an event $ e \in \mc{E} $.
    We assume that every event $ e \in \mc{E} $ has at least the following attributes: 
    $ \#_{\mi{act}}(e) \in \mc{A} $ assigns the activity to the event.
    $ \#_{\mi{time}}(e) $ assigns a timestamp to the event.
    $ \#_{\mi{case}}(e) $ assigns a case identifier to an event.
    $ \#_{user}(e) $ assigns the executing user to the event.
\end{definition}
\begin{definition}\label{def:basics3}
    \textbf{Trace}:
    A sequence $ \sigma = \seq{e_1,e_2,\dotsc e_n} $ of events is a trace, where $ \sigma(n) = e_n $ labels the event at index $ n $ inside the trace.
    For instance, \ $ \sigma(1) = e_1 $ is the first event of this sequence.
    All events in a trace have at least one attribute in common, the case identifier, so that
    $ \#_{\mi{case}}(\sigma(1)) = ... =  \#_{\mi{case}}(\sigma(n))$.
    A trace $ \sigma $ is ordered by the time attribute so that $ \forall e_n \in \sigma : \#_{\mi{time}}(e_n) < \#_{\mi{time}}(e_{n+1}) $.
\end{definition}
\begin{definition}\label{def:basics4}
    \textbf{Event Log}:
    Let $ \mc{T} $ be the universe of all traces and $ L \subseteq \mc{T} $ be an Event Log as a subset of all possible traces.
\end{definition}
\subsection{Gathering the HL Event Log}\label{subsec:gathering_high_level_log}
The required HL traces for the training phase are gathered by observing the ECS to obtain the actual user activities. Such an observer system was implemented accordingly.\footnote{\url{https://uni-ko.de/spm-observer}}
An important aspect here is that it is not necessary to know what LL events are triggered but only what user activity is possible (i.\,e., what functionality is provided by the ECS).
By defining triggers that record clicks on particular elements (e.g.,\,buttons), which represent certain HL activities, the HL events can be observed.
These events are also enriched with further attributes: the executing user, the timestamp, and the related workspace.
It is sufficient to do this \textit{only once} per system \textit{type} because other instances produce similar LLLs, which can be converted into HLLs by the already trained ECSEA model.

\Cref{tab:ecs_event_log_high_level} shows the recorded HLL of the related LL event log of \Cref{tab:ecs_event_log_excerpt}.\footnote{\textit{gws} means \textit{groupworkspace}}
Note that there is a relation but no explicit link between the LL and the HL traces (e.g.,\,the workspace as case identifier). In the following, both event logs are used as running examples and input for the training phase.
\begin{table}[H]
\centering
 \vspace*{-0.3cm}
\caption{Observed ECS HL log excerpt}\label{tab:ecs_event_log_high_level}
\scriptsize
\resizebox{1.0\columnwidth}{!}{
\begin{tabular}{|l|l|l|l|l|}
    \cline{1-5}
    \hcell{ID} & \hcell{USER\_UUID} & \hcell{C\_ID} & \hcell{TIMESTAMP} & \hcell{ACTIVITY} \\
    \cline{1-5}
    601 & 33d689884-ef3... & 1337 & 2021-11-11 10:47:16 & gws.filelibrary.file.created \\ 
    \cline{1-5}
    602 & 84f5d4ae2-434... & 1337 & 2021-11-11 10:47:18 & gws.wiki.created \\ 
    \cline{1-5}
    603 & 65788ec84-a12... & 1337 & 2021-11-11 10:51:11 & gws.wiki.wikiarticle.tag.created \\
    \cline{1-5}
    604 & 33d689884-ef3... & 1337 & 2021-11-11 10:56:44 & gws.wiki.wikiarticle.updated \\
    \cline{1-5}
\end{tabular}
}%
\end{table}
\subsection{Training of the Model}\label{subsec:ecs_event_abstraction_training}
Before starting the training algorithm, the data set is split into a training set (e.g.,\,80\%) and a test set (e.g.,\,20\%). 
The training set is used to train the model, and the test set is used to calculate the accuracy by applying the model to it to exclude overfitting. We define an ECSEA model in \Cref{def:model}.
\begin{definition}\label{def:model}
    \textbf{Model}: Let $ m = (\mi{llc},\mi{hlc}) $ be a model with two maps $ llc $ and $ hlc $.
    The map $ llc $ assigns single LL activities to a set of HL activities, i.e., 
    $ llc : \mc{A}_l \mapsto \{ \mc{X}\subseteq \mc{A}_h \} $.
    Further, let $ \mi{SLL} = \{ \seq{x_1, x_2, \ldots, x_n}| x_i \in \mc{A}_l\} $ the universe of sequences of LL activities. 
    The map $ \mi{hlc} $ assigns single HL activities to a set of sequences of LL activities that count their occurrences, i.e., $ hlc: \mc{A}_h \mapsto \{2^\mi{SLL},\mathbb{N}\} $
\end{definition}
To train the model, we use the training set of LL traces $ \mc{T}_l \subset \mc{T} $ and related HL traces $ \mc{T}_h \subset \mc{T} $.
We iterate over them and call for each valid combination a fitting function. The inputs for the function $ fit(t_l, t_h, \tau, \Gamma) $ are the traces $ t_l \in T_l $ and $ t_h \in T_h $ where $ \#_{case} (t_h(1)) = \#_{case} (t_l(1)) $.
The parameter $ \tau $ defines the maximal time-span in milliseconds between the timestamps of a LL event and a HL event.
$ \Gamma $ is a set of attribute names that are used to combine similar LL events and help to assign LL events to HL events.
Note that $ \Gamma $ can be an empty set $ \emptyset $.
Furthermore, all events in $ t_l $ and $ t_h $ should have all the attributes defined in $ \Gamma $, i.e., $ \forall e \in (t_l \cup t_h) : \nexists\ \gamma \in \Gamma\ s.t.\ \#_\gamma(e) = \varnothing $.
An example set for grouping attributes is $ \Gamma = \{ \mi{user} \} $, because it is obvious to only merge events from the same user.
Further attributes can be added to this set if both the LL events and the HL events have these attributes and they may be an indicator of a common HL activity in their same context.
The fitting function $ \mi{fit} $ now adjusts the model by adding or changing values in the maps $ hlc $ and $ llc $.

First, we create a set $ \mc{S} $ of sequences of all LL events where each event in a sequence has the same attributes of $ \Gamma $.
We build our sequences with:
\begin{align*}
\mc{S} = & \{\seq{ e_{1}, e_{2}, \dotsc, e_{n} }: e_{i} \in t_l\, | \\
& \#_\mi{time}(e_{i}) < \#_\mi{time}(e_{i+1}) \land\,\forall \gamma \in \Gamma:\,\#_\gamma(e_{i}) = \#_\gamma(e_{i+1}) \}
\end{align*}

\begin{example}\label{ex:sub-sequences}
Considering the LL trace from \Cref{tab:ecs_event_log_excerpt} and a grouping attribute set $ \Gamma = \{ user \} $, we create the following sequences of events (using the ID column as identification of the event):
{\scriptsize
\begin{align*}
 \mc{S} = \{ \seq{ 104, 105, 107, 109, 112 }, \seq{ 106, 108 }, \seq{ 110, 111 } \} 
\end{align*}}%
\end{example}
\noindent Next, we create a mapping $ \mc{M} $ for each HL event $ e_h \in t_h $.
For this, we define a function $ \mc{S}_h(e_h) $ that returns for $ e_h $ the related sub-sequence $ s \in \mc{S}\ :\ \forall \gamma \in \Gamma\ : \#_{\gamma}(e_h) = \#_{\gamma}(s(1)) $.
Then, we create new sub-sequences with the events in $ s $ where the time distance between the time attribute of the LL events and the HL event $ e_h $ is below the time-span variable $ \tau $.
Furthermore, if a LL event can be assigned to multiple HL events, only the HL event with the minimal temporal distance is selected for the mapping.
So we create our mapping with:
\begin{align*}
\mc{M} =\ &\{ e_h \in t_h \mapsto \seq{ e_{1}, e_{2}, \dotsc, e_{n} }\ : e_{i} \in \mc{S}_h(e_h)\ |\\
 & \#_\mi{time}(e_{i}) < \#_\mi{time}(e_{i+1})\, \land\, \\ 
 & |\#_{\mi{time}}(e_h) - \#_{\mi{time}}(e_i)| \leq \tau\ \land\\
 & \nexists\ e_{h2} \in t_h : ( |\#_{\mi{time}}(e_{h2}) - \#_{\mi{time}}(e_i)| < \\
& \ \ |\#_{\mi{time}}(e_h) - \#_{\mi{time}}(e_i)|\ \land\ \mc{S}_h(e_h) = \mc{S}_h(e_{h2})\ ) \ \}
\end{align*}
\begin{example}\label{ex:mapping}
Considering the HLL in \Cref{tab:ecs_event_log_high_level}, the sub-sequences from \Cref{ex:sub-sequences} and the value $ 5s $ for $ \tau $ we get the following mapping $ \mc{M} $.
{\scriptsize
\begin{align*}
M\,=\,\{\,601:\,\seq{104,105,107},\,602:\,\seq{ 106,108},\,603:\,\seq{ 110,111},\,604: \seq{ 112}\,\}
\end{align*}}%
\end{example}

\noindent Next, we create a map $ \mc{M}^\prime $ by extracting all activities from the LL and HL events in $ \mc{M} $ and convert the found event sequences into activity sequences.
\begin{align*}
 \mc{M}^\prime =\ & \{ \#_\mi{act}(e_h) : e_h \in \mc{M} \mapsto \\
 & \seq{ \#_\mi{act}(e_1), \#_\mi{act}(e_2), \dotsc, \#_\mi{act}(e_n)}\ : e_n \in \mc{M}(e_h)\} 
\end{align*}

\begin{example}\label{ex:a_mapping}
With $ \mc{M} $ from \Cref{ex:mapping} we get:
{\scriptsize
\begin{alignat*}{2}
 \mc{M}^\prime =\ & \{  \mi{gws.filelibrary.file.created} : \\
    &\quad \seq{ \mi{file.file.created},\mi{file.collection.file.added},\\ 
    &\quad \quad \mi{files.file.notification.set} }, \\
 &\mi{gws.wiki.created} : &\\
 &\quad \seq{ \mi{wiki.page.created}, \mi{wiki.page.follow} }, \\
 &\mi{gws.wiki.wikiarticle.tag.created} : \\
 &\quad \seq{ \mi{wiki.page.updated}, \mi{wiki.page.tag.added} }, \\
 &\mi{gws.wiki.wikiarticle.updated} : \\
 &\quad \seq{ \mi{wiki.page.updated} }\} &&
\end{alignat*}}%
\end{example}

\noindent Finally, we add the mappings in $ \mc{M}^\prime $ to the $ hlc $ map.
If a LL activity sequence already exists in this mapping, we increment the counter value for this sequence.
\begin{example}
With $ \mc{M}^\prime $ from \Cref{ex:a_mapping} and a previous empty $ \mi{hlc} $ map we get:
{\scriptsize
\begin{alignat*}{2}
 \mi{hlc} =\ & \{
 \mi{gws.filelibrary.file.created} : & \\
 &\quad \{ \seq{ \mi{file.file.created},\mi{file.collection.file.added} ,\\ 
 & \quad\quad\mi{files.file.notification.set} }, 1 \}, \\
 &\mi{gws.wiki.created} : \\
 &\quad\{\seq{ \mi{wiki.page.created}, \mi{wiki.page.follow} }, 1 \}, \\
 &\mi{gws.wiki.wikiarticle.tag.created} : \\
 &\quad\{\seq{ \mi{wiki.page.updated}, \mi{wiki.page.tag.added} }, 1\}, \\
 &\mi{gws.wiki.wikiarticle.updated} : \\
 &\quad\{\seq{ \mi{wiki.page.updated}}, 1\}  \} 
\end{alignat*}}%
\end{example}

\noindent Furthermore, we update the $ \mi{llc} $ map according to the reverse mapping of $ \mc{M}^\prime $ and add all found possible LL activity to HL activity mapping to this list.

\begin{example}
With $ \mc{M}^\prime $ from \Cref{ex:a_mapping} and a previous empty $ \mi{llc} $ map we get:
{\scriptsize
\begin{alignat*}{3}
 \mi{llc} =\ & \{
 \mi{file.file.created}            : &\quad&\{ \mi{gws.filelibrary.file.created} \} \\
 &\mi{file.collection.file.added}   : &\quad&\{ \mi{gws.filelibrary.file.created} \} \\
 &\mi{files.file.notification.set}  : &\quad&\{ \mi{gws.filelibrary.file.created} \} \\
 &\mi{wiki.page.created}            : &\quad&\{ \mi{gws.wiki.created} \} \\
 &\mi{wiki.page.follow}             : &\quad&\{ \mi{gws.wiki.created} \} \\
 &\mi{wiki.page.tag.added}          : &\quad&\{ \mi{gws.wiki.wikiarticle.updated},\\ &&&\quad\mi{gws.wiki.wikiarticle.tag.created} \} \\
 &\mi{wiki.page.updated}            : &\quad&\{ \mi{gws.wiki.wikiarticle.updated} \} \} &&
\end{alignat*}}%
\end{example}

\noindent This was one call of the $ \mi{fit} $ function with a LL trace and its related HL trace.
By iterating over all traces, we create a model that knows which HL activities are associated with which possible LL sequences.
The more traces are used to fit the model, the more combinations may be added to the maps $ \mi{hlc} $ and $ \mi{llc} $.
Note that because no HL event was in the temporal range of event 109, this event was ignored as a ghost event and no mapping was created for the LL activity  $ \mi{community.visit}$. 
However, the final map may contain sequences that are incorrect.
For instance, a LL activity may be mapped to the wrong HL activity.
Thus, we try to ignore these wrong mappings in the application phase (\Cref{subsec:ecs_event_abstraction_application}) by excluding these mappings by calculating an error score.

To evaluate the model, we apply it to the train data set and calculate its accuracy.
We iterate over the LL traces of the train data set and call the second algorithm (cf. \Cref{subsec:ecs_event_abstraction_application}) that converts a LL trace into a HL trace.
Then, we compare the generated traces and the related \textit{original} HL traces from the train data set and compute the accuracy.
We use a normalized distance similarity to calculate the difference between two activity sequences.
Hence, a minimal difference between the sequences (e.g., a swap of two activities) has no major influence on the accuracy.
We use the Damerau–Levenshtein distance \cite{Damerau1964} ($ dld $) to calculate this distance similarity.
The distance calculates how many deletions, insertions, or swaps are required to convert one sequence into another one.
The normalized distance similarity is a value between 0 and 1 and defines how similar two sequences are while considering their length.
So we get our accuracy with
\begin{align*}
    \mi{acc}(train) = 
    \dfrac{\sum\limits_{t_p,t_h\in \mi{train}}1-{\mi{dld}(t_p,t_h)} \bigg/ {\mi{max}(|t_p|,|t_h|)}}
    {|\mi{train}|}
\end{align*}
\noindent
The training of the model is conducted several times until the accuracy is maximal. 
In this context, the accuracy depends on the values of the hyperparameters $\tau$, $ \Phi $, and $ \vartheta $ (see \ref{subsec:ecs_event_abstraction_application}). 
Hence, we choose possible different values of $\tau$, $ \Phi $, and $ \vartheta $ and perform the training and evaluation in the sense of a grid search-based hyperparameter optimization.
The model with the best accuracy is finally chosen, and its accuracy is evaluated again against the test set to assess its degree of overfitting.
\subsection{Applying the Model}\label{subsec:ecs_event_abstraction_application}
A greedy algorithm based on sliding windows converts LL traces into HL traces using the ECSEA model.
\def\lineref#1{(line \ref{#1})}
\begin{algorithm}[caption={Model Application},label={alg:application}]
function $ \mi{apply}(m, t_l, \vartheta, \Phi, \tau, \Gamma)$:
  $ \mi{lastlen} \leftarrow t_l.len() + 1, \mi{\beta} \leftarrow [\,]$ §\label{line:defLastLenBeta}§
  while $ t_l.len() > 0 $ do: §\label{line:loopUntilEmpty}§
    if $ t_l.len() = \mi{lastlen} $: §\label{line:checkLenChanged}§
      $ t_l \leftarrow t_l.removeFirst() $
      $ \mi{continue} $
    $ \mi{lastlen} \leftarrow t_l.len() $
    $ \mc{W} \leftarrow \mi{getFirstWindow}(t_l, \tau, \Gamma) $ §\label{line:create_windows_event_list}§
    $ \psi, a_h \leftarrow \mi{getBestMapping}(m, \mc{W}, \vartheta) $ §\label{line:get_best_mapping}§
    if $ \psi = \varnothing $:
      $ \mi{continue} $ §\label{line:continue_when_no_bestMapping}§
    $ \beta.\mi{append(createEvent(}\mc{W},\psi,a_h,\Phi))$ §\label{line:createEvent}§
    $ t_l \leftarrow \mi{removeEvents}(t_l, \mc{W}, \psi) $ §\label{line:shrinkTl}§
  $ \beta \leftarrow \mi{mergeEvents}(\beta, \Gamma, \Phi, \tau) $ §\label{line:merge_events}§
  return $ \mi{createTrace}(\beta) $ §\label{line:create_trace}§
\end{algorithm}

The function $ \mi{apply}(m, t_l, \vartheta, \Phi, \tau, \Gamma)$ converts the LL trace $ t_l $ with the model $ m $ into a HL trace.
We use the mapping threshold parameter $ \vartheta $ to accept or decline possible mappings.
The enum parameter $ \Phi $ (\textit{timestamp-merge-type}) is used to define how the timestamp attributes of the LL events are merged for the new HL timestamp attribute.
Valid values are $ \mi{MIN} $ (for the lowest timestamp value), $ \mi{MAX}$ (for the highest timestamp value), $ \mi{MEAN} $ (for the mean value of all timestamps), and $ \mi{MEDIAN} $ (for the median value of all timestamps).
Furthermore, we use the parameter $ \tau $ and $ \Gamma $, which we also use in the training phase. 
This algorithm is illustrated in \Cref{alg:application}.

In the first step \lineref{line:defLastLenBeta} we create a variable $ \mi{lastlen} $ with the value of the length of the LL trace $t_l$ (+1) as we need it later in the loop. 
Furthermore, we initialize an empty list $ \mi{\beta} $, that will contain the newly created HL events.
Then we loop until the sequence $t_l$ of LL events is empty \lineref{line:loopUntilEmpty}.
In the loop, we first check if the last iteration did not change the LL trace \lineref{line:checkLenChanged}. 
If this is the case, we remove the first event in $t_l$ to eliminate events, that are ghost events (for handling \textbf{C5}) and continue then with the next iteration.
Else, we store the length of the current $t_l$ in $ \mi{lastlen} $.
Now, we calculate the first window $ \mc{W} $ from the trace $ t_l $ where the events of the window have all the same attribute values defined in $ \Gamma $ (for supporting \textbf{C2 \& C4}) and where the timestamp attributes of the first and last event have the maximal temporal distance of $ \tau $. The first event in this window is always the first event of the current $ t_l $.
So, we get our window with:
\begin{align*}
 \mc{W} =\  &\seq{e_1,\dotsc,e_n},\ e_1 = t_{l}(1)\ \land\\
  & \,e_i \in t_l : \#_\mi{time}(e_i) < \#_\mi{time}(e_{i+1})\ \land\\
            & \,\forall \gamma \in \Gamma\,:\,\#_\gamma(e_{i}) = \#_\gamma(e_{i+1})\ \land\\
            & \,(\#_{\mi{time}}(e_{i}) - \#_\mi{time}(e_{1})) \leq \tau 
\end{align*}

For this window, we try to get the best possible mapping $ \psi $ with the related HL activity $ a_h $ \lineref{line:get_best_mapping}.
We conduct a greedy search for the local optimum based on the assumption that the model contains the mappings that gives us the best solution for the current window.
Thus, the central part in this algorithm is the function $ getBestMapping $, which is defined in \Cref{alg:get_best_mapping}. 
This function uses the information from the model $ m $ and the mapping threshold $ \vartheta $ to create a sequence of activities from the window $ \mc{W} $. 
We name this list of events the best mapping $ \psi $.
Furthermore, this function returns the related HL activity $ a_h \in A_h $.
The idea is that we first get all possible HL activities (for handling \textbf{C4}) based on the LL activities in the window $ w $, which are stored in the $ llc $ map of the model \lineref{line:possibleMappings}.
Then, we iterate over these HL activities \lineref{line:iteratehla} and iterate again over their mappings and their counting number from the $ hlc $ map \lineref{line:iteratehlc}.
If all LL activities of this mapping are present in the LL activities of the window \lineref{line:ifallactivitiesinwindow}, we calculate an error score \lineref{line:calcerrorscore}.
This error score uses the Damerau–Levenshtein edit distance \cite{Damerau1964} (for handling \textbf{C3 \& C4}), which we normalize by dividing through the maximum size of both sequences.
Furthermore, this error score is influenced by the counter variable, which expresses the number of times this mapping was found in the training phase.
If we found a better error score than the previous ones \lineref{line:ifbetterscore}, we save the best error, the mapping in the $ \psi $ variable, and the current HL activity $ \mi{hla} $ in a temporal variable.
After iterating over all mappings, we can return the best mapping $ \psi $ and the related HL activity $ a_h $ \lineref{line:returnpsi}.

We now continue with the main function $ \mi{apply} $ (Algorithm\ \ref{alg:application}).
If the function $ getBestMapping $ returns no mapping we continue the loop with the next window \lineref{line:continue_when_no_bestMapping}, which then starts one event later.
Otherwise, if there is a mapping, we call the function $ createEvent $ and append the result to $ \beta $ \lineref{line:createEvent}.
This function merges all events inside the window $ \mc{W} $ with the mapped activities $ \psi $ into a new HL event with the HL activity $ a_h $ (for handling \textbf{C1}). 
The timestamp of this new event is defined by the enum parameter $ \Phi $. 
We assign a new timestamp attribute to the new event: the lowest/highest timestamp value from the events if $ \Phi = \mi{MIN} $ or $ \Phi = \mi{MAX} $, or the mean/median of all timestamp values from the events if $  \Phi = \mi{MEAN} $ or $  \Phi = \mi{MEDIAN} $, respectively.
All attributes defined in $ \Gamma $ can also be assigned to this new event, as all events have equal values for these attributes.
E.\,g., the user attribute of the LL events can be assigned to the user attribute of the new HL event.
Finally, we shrink $ t_l $ and remove all events from the window $ \mc{W} $ which were \lineref{line:shrinkTl} merged to the new HL event.

After all windows have been processed, we have a list of HL events in the variable $ \beta $.
It is possible that there are events that a) are assigned to the same activity, b) have the same attribute values for $ \Gamma $, and c) their timestamp distance is below $ \tau $. 
If we find events that fulfill these three conditions, we merge them to one HL event with the function $ mergeEvents $ \lineref{line:merge_events} and calculate the timestamp attribute like before with $ \Phi $.
Now, the resulting list $ \beta $ contains the newly created HL events that can be converted into a HL trace \lineref{line:create_trace}.

\begin{figure}[H]
\begin{algorithm}[caption={Get Best Mapping},label={alg:get_best_mapping}]
function $ \mi{getBestMapping}(m, w, \vartheta)$:
  $ \mi{possibleHighLevelActivities} \leftarrow \mi{getPossibleHighLevelAct(w, m.llc)} $ §\label{line:possibleMappings}§
  $ \psi \leftarrow \varnothing, a_h \leftarrow \varnothing,  \mi{bestError} \leftarrow \infty $ 
  forall $ \mi{hla} \in \mi{possibleHighLevelActivities} $ do: §\label{line:iteratehla}§
    forall $ \mi{mapping}, \mi{counter} \in \mi{m.hlc(hla)} $ do: §\label{line:iteratehlc}§
      if $ \forall\,\mi{lla}_1 \in \mi mapping: \exists\,\mi{lla}_2 \in w : \mi{lla}_2 = \mi{lla}_1 $: §\label{line:ifallactivitiesinwindow}§
         $ \mi{error} \leftarrow \frac{\mi{dld}(w, \mi{mapping})}{max(|w|,|mapping|)\ \cdot\ \sqrt{\mi{counter}}}$ §\label{line:calcerrorscore}§
        if $ \mi{error} < \mi{bestError} \land \mi{error} < \vartheta$:§\label{line:ifbetterscore}§
          $ \mi{bestError} \leftarrow  \mi{error}, \psi \leftarrow \mi{mapping}, a_h \leftarrow \mi{hla} $
  return $ \psi, a_h $§\label{line:returnpsi}§
\end{algorithm}
\vspace*{-1cm}
\end{figure}

\section{Evaluation}\label{sec:evaluation}
We implemented ECSEA\footnote{ECSEA framework: \url{https://uni-ko.de/spm-ecsea}} and evaluated it.
As a first pre-check, we applied ECSEA to a real-life HLL with synthetically generated LLLs (\ref{subsec:evaluation_artifical_data}). 
With the confirmation that the algorithm works, we observed a HLL of an ECS for three months and afterwards applied ECSEA with the corresponding LLL to it (\ref{subsec:evaluation_real_ecs_data}).

\subsection{Instantiation with Synthetic Data}\label{subsec:evaluation_artifical_data}
For a first evaluation,\footnote{Experiment setup: \url{https://uni-ko.de/ecsea-testing}} we used the event log \textit{PermitLog} from the BPI Challenge 2020.\footnote{BPI Challenge 2020: \url{https://icpmconference.org/2020/bpi-challenge}}
We assume that the log is a HLL, so we have to create the LLLs synthetically.
First, we split each activity of the original log into multiple LL activities and generated synthetic LLLs with different configurations.
Thereby, the created LLLs met the characteristics mentioned in \Cref{subsec:ecs_event_log_characteristics}.
The \textit{number of new LL activities} defines how many synthetic LL activities should be created for each HL activity.
Thus, we created for each HL event multiple LL events and assigned them different (nearby) timestamps.
Furthermore, we randomly sampled 10\% of the traces in the training phase to demonstrate that only a small part of the homogeneous event log is needed to train a model that is able to abstract the remaining part of the LLL.
For each configuration, we created ten different LLLs and calculated the average of the results for the final charts.
In total, we generated 70 LLLs. 
Then, we used the training algorithm to train multiple ECSEA models for each of these LLLs. 
For this, we used different parameter values for the \textit{timestamp-merge-type} $ \Phi $, the default value for the \textit{mapping threshold} $ \vartheta $, and used a fixed value for the \textit{maximal time-span} $ \tau $ based on the parameter \textit{number of LL activities}.
Furthermore, we set the parameter $ \Gamma\,=\,\{ \mi{org{:}resource, org{:}role} \} $ as these attributes are present in the LL and HL events and thus can be used to find related LL events.
In total, we applied the algorithms 280 times. 
Finally, we could use the different training parameters, the test accuracy, and the running times to plot charts. 
\begin{figure}[t]
    \centering
    \begin{subfigure}[b]{0.495\textwidth}
        \centering
        \includegraphics[width=\textwidth]{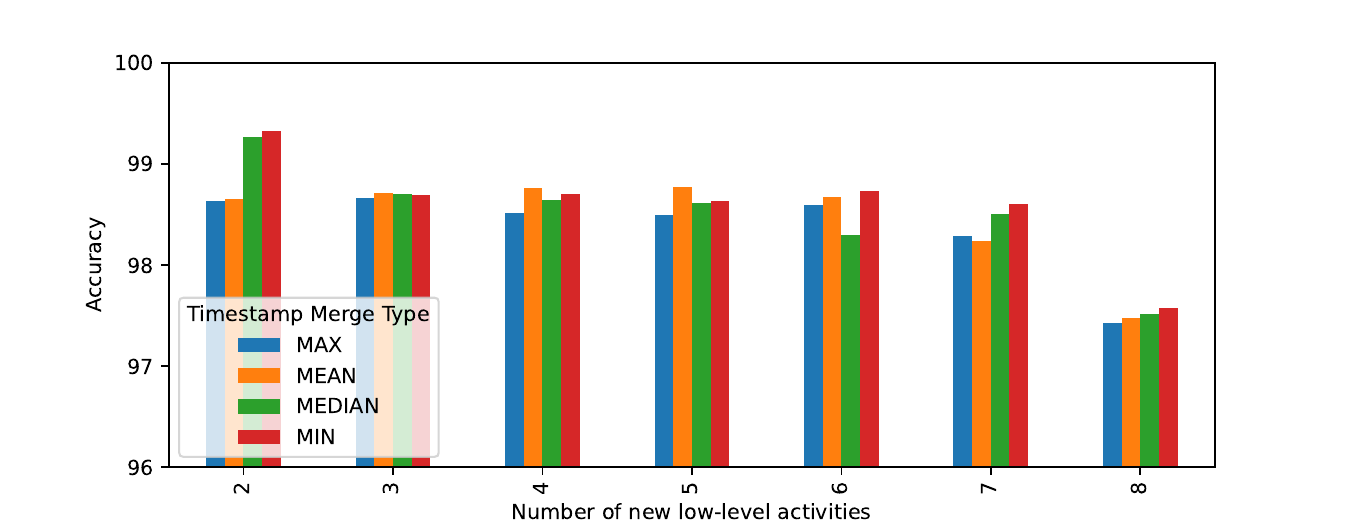}
        \caption{Test accuracy}
        \label{fig:eval_test_accuracy}
    \end{subfigure}
    \begin{subfigure}[b]{0.495\textwidth}
        \centering
        \includegraphics[width=\textwidth]{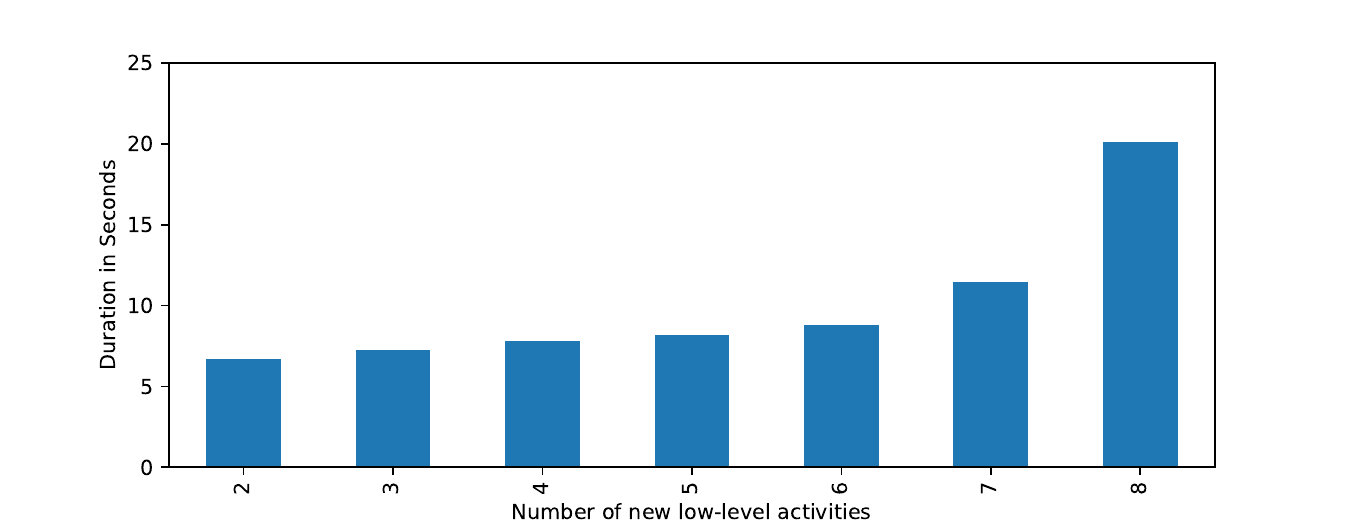}
        \caption{Train duration}
        \label{fig:eval_train_duration}
    \end{subfigure}
    \caption{Evaluation results of the artificial LL event data}
    \label{fig:evaluation_results_artificial_data}
    \vspace*{-0.5cm}
\end{figure}
We ran our experiments on Ubuntu 22.04LTS, using 8 CPU Cores and 32GB memory. 
We used 706 of the 7065 traces (randomly sampled) to train the model.
The original HLL contains 86581 events (training set: 8534 events) and 45 HL activities.
The generated LLLs with two LL activities per HL activity consist of about 180k events (training set about 17.8k events), and about 117 LL activities.
The generated LLLs with eight LL activities per HL activity consist of about 700k events (training set about 69k events), and about 400 LL activities.

\Cref{fig:eval_test_accuracy} shows the accuracy for the test data sets.
The bars represent the \textit{number of new LL activities} and the parameter \textit{timestamp-merge-type} $ \Phi $.
The accuracy is always above 98\% and drops slightly the more LL activities are used per HL activity.
The \textit{timestamp-merge-type} $ \Phi $ influences the accuracy only slightly.
\Cref{fig:eval_train_duration} shows the duration of the training phase.
The more LL activities are generated, the longer the duration. 
The results of the first evaluation show that ECSEA works successfully with \textbf{synthetic data} and that the characteristics from \cref{subsec:ecs_event_log_characteristics} can be handled by the algorithm.
\subsection{Instantiation with Real ECS Data}\label{subsec:evaluation_real_ecs_data}
The second evaluation was performed with \textbf{real-world data} using the event log of UniConnect, an operational large-scale Enterprise Collaboration System with more than 3000 users. 
We generated the HLL (containing activity, timestamp, workspace, executing user) with the help of our observer for a period of three months. 
We then extracted the LL events from the ECS for the same time period and converted both logs into XES using the workspace as case id. 
We trained an ECSEA model from the input data.
As we have in both logs the \textit{user} attribute, we can use this attribute name in the $ \Gamma $ set for the detection of LL events that possibly can be mapped to HL events.
Furthermore, we used hyperparameter optimization as described above to find the best value for the parameters \textit{mapping threshold} $ \vartheta $ and \textit{maximal time-span} $ \tau $.
Finally, we reached an accuracy up to 96\%. 
We are now able to abstract the entire LLL of UniConnect (containing the last eight years), which will give us an extensive data basis for SPM.\footnote{For privacy reasons, we cannot publish the data set.}
To demonstrate that the ECSEA model is viable for other instances of the same ECS, we evaluated it further and (successfully) applied it to KoCo (another instance of HCL Connections).

\section{Discussion and Outlook}\label{sec:discussion_outlook}
In this paper, we have developed and evaluated a novel EA approach that considers the special characteristics of ECS LLLs.
The baseline algorithm of ECSEA learns an EA model by means of an iterative comparison of observed HL traces and native LL traces.
With this model, it is possible to abstract an ECS HLL from ECS LLLs.
The trained model can also be used to abstract LLL from different instances of the same system type (without major customizations), resulting in a high reusability of the trained model in research and practice. 
The precondition is that a HLL must be recorded through observation which is a limitation, however, this has to be done only once. 
In this paper, we have demonstrated the feasibility, performance, and accuracy of ECSEA using synthetic and real-world logs.
The evaluation results show that ECSEA works and produces accurate HLLs. 
Further preprocessing is now needed for the HLL to tackle the challenge of finding a suitable case in the abstracted event log. 
Until now, we only used the workspace as a case identifier. 
However, to identify collaboration patterns, we need to reveal the nucleus of the collaboration, which can then serve as the case id in the next step. 
As a first candidate, we will consider the set of social documents that is jointly worked on.

Although ECSEA was designed to address the characteristics of ECS  logs, it can be applied in other contexts as well where HLLs are observable. 
This is made possible by the increasingly widespread browser-based information systems where an observer can be injected.

A possible extension of the algorithm can create two HL events with the \textit{min} and \textit{max} timestamp from the selected LL events as \textit{activity instance} with a \textit{start} and \textit{end} life-cycle transaction respectively.
Furthermore, we may extend the baseline algorithm and add weight factors for the $ hlc $ mappings, which we can optimize with a genetic algorithm.

\section*{Acknowledgments}
This work was partly funded by the Deutsche Forschungsgemeinschaft (DFG) – project number 445182359.

\bibliographystyle{IEEEtran}
\bibliography{main}
%

\end{document}